\documentclass[journal]{IEEEtran}
\usepackage{cite}
\usepackage{gensymb}
\usepackage[caption=false,font=normalsize,labelfont=sf,textfont=sf]{subfig}
\usepackage{amsmath,amssymb,amsfonts,bm}
\usepackage{algorithmic}
\usepackage{graphicx}
\usepackage{textcomp}
\usepackage{xcolor}
\usepackage{optidef}
\usepackage{url}  
\usepackage{booktabs}
\usepackage[linesnumbered,ruled,vlined]{algorithm2e}
\usepackage[export]{adjustbox}
\usepackage{cleveref}
\usepackage{upgreek}
\usepackage{flushend,cuted}
\usepackage{textcomp}
\definecolor{codegray}{gray}{0.95}

\crefformat{section}{\S#2#1#3} 
\crefformat{subsection}{\S#2#1#3}
\crefformat{subsubsection}{\S#2#1#3}

\newtheorem{definition}{Definition}

\begin{document}
% at, but, by, for, in, nor, of, on, or, the, to, and up, which are usually
\title{Fusion Intelligence for Digital Twinning AI Data Centers: A Synergistic GenAI-PhyAI Approach}

% \author{
% Authors*
% % Zhiwei Cao,
% % Yonggang Wen,~\IEEEmembership{Fellow,~IEEE,}
% % <-this % stops a space
% }
% \markboth{Journal of \LaTeX\ Class Files,~Vol.~14, No.~8, August~2015}%
% {Shell \MakeLowercase{\textit{et al.}}: Bare Demo of IEEEtran.cls for IEEE Journals}
\author{
Ruihang~Wang,~Minghao~Li,~Zhiwei~Cao,~Jimin~Jia,~Kyle~Guan,~and~Yonggang~Wen*,~\IEEEmembership{Fellow,~IEEE}%
 \thanks{
 Authors are with the College of Computing and Data Science, Nanyang Technological University (NTU), Singapore (e-mail: ruihang.wang@ntu.edu.sg; minghao002@e.ntu.edu.sg; zhiwei003@e.ntu.edu.sg; jimin.jia@ntu.edu.sg; ygwen@ntu.edu.sg).
}%
 \thanks{Kyle Guan is an independent researcher (e-mail: kcguan@gmail.com).}%
\thanks{*Corresponding author}

}

% \vspace{-10mm}

\maketitle
\begin{abstract}
The explosion in artificial intelligence (AI) applications is pushing the development of AI-dedicated data centers (AIDCs), creating management challenges that traditional methods and standalone AI solutions struggle to address. While digital twins are beneficial for AI-based design validation and operational optimization, current AI methods for their creation face limitations. Specifically, physical AI (PhyAI) aims to capture the underlying physical laws, which demands extensive, case-specific customization, and generative AI (GenAI) can produce inaccurate or hallucinated results. We propose Fusion Intelligence, a novel framework synergizing GenAI's automation with PhyAI's domain grounding. In this dual-agent collaboration, GenAI interprets natural language prompts to generate tokenized AIDC digital twins. Subsequently, PhyAI optimizes these generated twins by enforcing physical constraints and assimilating real-time data. Case studies demonstrate the advantages of our framework in automating the creation and validation of AIDC digital twins. These twins deliver predictive analytics to support power usage effectiveness (PUE) optimization in the design stage. With operational data collected, the digital twin accuracy is further improved compared with pure physics-based models developed by human experts. Fusion Intelligence offers a promising pathway to accelerate digital transformation. It enables more reliable and efficient AI-driven digital transformation for a broad range of mission-critical infrastructures.
\end{abstract}

% Note that keywords are not normally used for peer-reviewed papers.

\IEEEpeerreviewmaketitle
\section{Introduction}
\begin{figure}[t]
    \centering
    \includegraphics[width=1\linewidth]{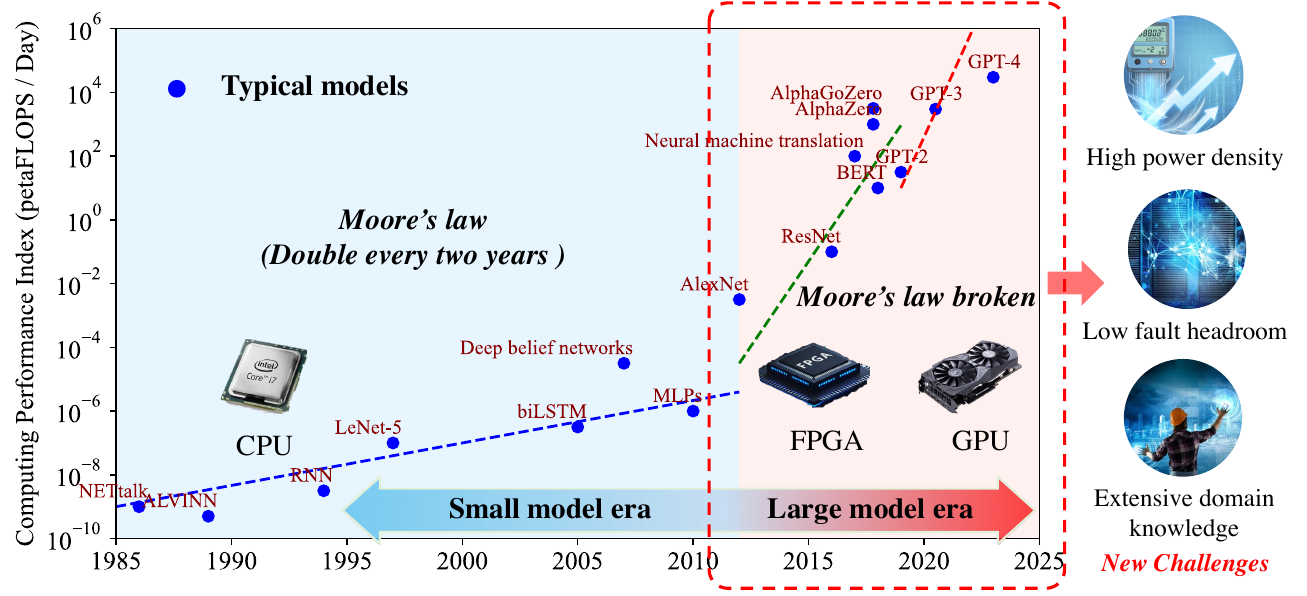}
    \caption{Growth of computational performance demand over the past four decades. Moore’s law is broken after entering the large model era~\cite{zhu2023intelligent}. Such growth introduces new challenges for AIDC infrastructure management.}
    \vspace{-5mm}
    \label{fig:computility}
\end{figure}

% {\textcolor{red}{[\textbf{Background}]
Mission-critical infrastructures (MCIs) are the socioeconomic backbone to ensure public welfare and safety. They span sectors such as agriculture, building, communication, data centers, energy systems, and fleets. Artificial intelligence-dedicated data centers (AIDCs) are one of the MCIs that provide computing capability for massive data processing in the large AI model era. They rely on heterogeneous parallel computing hardware, such as GPUs and FPGAs, to meet the demand of growing computing performance as shown in Fig.~\ref{fig:computility}. According to the IDC report~\cite{global-computing-index}, an increase in the national computing capability index will stimulate the growth of 3.6\perthousand{} in the nation's digital economy and 1.7\perthousand{} in Gross Domestic Product (GDP). Therefore, the development of AIDC is essential for driving digital economic growth. 

% \textcolor{red}{[\textbf{Challenges}]
However, the emerging AIDC presents three management challenges, i.e., short cooling response time, low fault headroom, and talent shortage. Firstly, the high per-rack power density of AIDC amplifies operational risks~\cite{aidc-rack-power}. As the power density of each rack escalates, fluctuations in dynamic IT workload shorten the response time of the cooling systems, in turn increasing the risk of server overheating. Secondly, high resource utilization in AIDC reduces fault tolerance. Given the elevated construction and operational costs of these infrastructures, AIDC often maintains GPU resource utilization rates above 85\% to ensure operational profitability~\cite{ai-infra}, in turn resulting in a decrease in system headroom for fault tolerance. Thirdly, managing an AIDC requires the integration of cross-domain knowledge, including the computing, cooling, and power supply systems. Nevertheless, 53\% of DC operators encounter challenges in recruiting qualified professionals.

% \textcolor{red}{[\textbf{Common Efforts and Gaps}]
Facing these challenges, the AIDC industry urgently calls for intelligent solutions, as human capabilities alone are insufficient to address them. Currently, most DCs are equipped with sensing, computing, and control systems for reliable operations. To achieve intelligent management, the computing system can adopt advanced machine learning (ML) algorithms for analyzing data and deriving control actions. For example, deep reinforcement learning (DRL) has been applied for safe and energy-efficient DC cooling~\cite{li2019transforming} or joint IT-facility~\cite{zhou2021joint} control optimizations. However, these ML techniques face two gaps in real-world deployment. {\em The first is the data availability issue}. Pure data-driven ML models usually require a significant amount of data, including the abnormal cases that are difficult to collect in a stably operated system. Physically collecting these data may generate high costs. ML models are susceptible to overfitting, they could produce physically implausible predictions when evaluated on data points that deviate significantly from the training data. {\em The second is the black-box problem}. The risk-aversion mindset in DC industry curtails implementing ML-based policies, as they may introduce uncertainties and risks without adequate safety validation.

% \textcolor{red}{[\textbf{Latest Efforts and Gaps}]
The digital twins have been touted as a generalized concept to release the potential of ML by providing a virtual environment for validating such intelligent policies~\cite{wang2020kalibre}. AIDC digital twins consist of the virtual {\em 3D scenes}, the dynamically updated {\em mechanistic models}, and the evolving set of {\em data} once the system is in operation. The 3D scenes include the layout of an environment with a set of simulation-ready assets~\cite{SimReadyAssets}, such as racks, computer room air handling units (CRAHs), cooling distribution units (CDUs), chillers, etc. On top of the digital scenes, the mechanistic models need to simulate the system dynamics and synchronize with the physically collected data. To create digital twins, generative AI (GenAI), such as large language models (LLM), has been adopted to generate the scene configurations or simulation codes based on user-specified prompts~\cite{feng2023layoutgpt}. For example, a user may want to design an AIDC with a 1MW IT capacity and a design power usage effectiveness (PUE) below 1.3. GenAI can analyze the user’s objective and generate digital twins with the 3D scenes and mechanism simulation codes to meet these requirements. However, a critical limitation of GenAI is its susceptibility to {\em hallucinations}, such as generating infeasible layouts (e.g., rack placements that violate certain standards) or misinterpreting user objectives. To accurately perceive, understand, and perform complex actions in the physical system, the emerging physical AI (PhyAI) is proposed by integrating technologies such as physics-informed machine learning (PIML)~\cite{karniadakis2021physics} and DRL~\cite{physicalAI}. PhyAI is advantageous in developing the mechanistic models with improved sample efficiency and physical plausibility by incorporating physical laws. However, creating these models requires both domain-specific knowledge (e.g., expert-curated loss functions) and fine-tuning processes. This case-by-case tuning hinders scalability for large-scale deployment.  

\begin{figure*}[t]
    \centering
    \includegraphics[width=1\linewidth]{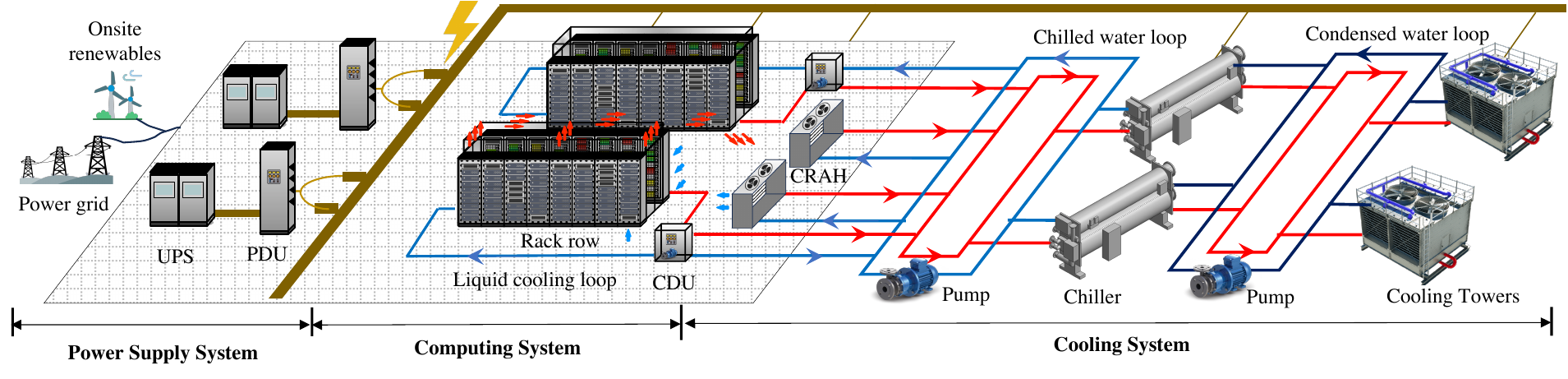}
    \caption{The overview of the AIDC system that consists of the power supply, computing, and cooling systems. Compared to traditional DC, AIDC is usually equipped with a liquid cooling system to remove the heat generated by high-power-density racks. The interconnected system presents multiple modeling challenges such as large-scale layout construction, data availability, data quality, multiphysics, and multiscale nature.}
    \label{fig:aidc}
\end{figure*}

% \textcolor{red}{[\textbf{Our Proposal}]
To overcome the limitations of relying solely on GenAI and PhyAI, we propose a {\bf fusion intelligence} framework to synergize their respective unique capabilities. Specifically, the fusion intelligence integrates GenAI’s pretrained broad knowledge (e.g., through LLM's self-supervised learning) with PhyAI’s knowledge requirements (e.g., PINN-enforced conservation laws) to develop physically plausible digital scenes and mechanistic models. This fusion creates a closed-loop collaboration where GenAI and PhyAI interact synergistically. Specifically, in the forward path, the GenAI first leverages its broad domain knowledge to define the semantic digital scenes, model structures, variables, or relationships that align with specific AIDC requirements, while PhyAI enforces these physical constraints during model parameter learning. In the backward path, the PhyAI’s outputs inform GenAI's subsequent generations. If the model output deviates from the target (e.g., online real-time data or certain constraints), GenAI reflects on the results and refines the generation. This dual-loop interaction reduces GenAI's hallucinations and PhyAI's modeling efforts, enabling automated, reliable, and scalable digital transformation for evolving AIDC infrastructures.

% \textcolor{red}{[\textbf{Case Studies}]{
The fusion intelligence framework demonstrates its efficacy in two typical AIDC applications.
In physics-aware design generation, the framework automates equipment selection for designing an energy-efficient AIDC with 50-petaFLOPS computing performance. For generative coil heat exchanger modeling, fusion intelligence achieves a lower prediction error compared to the pure physics-based model developed by a human expert. Both cases underscore fusion intelligence’s ability to harmonize automation, physical plausibility, and scalability in AIDC lifecycle management. The AIDC case study provides insights into transforming MCI management toward a highly automated and efficient level.

\section{AIDC Overview \& Modeling Challenges}
This section provides an overview of the interconnected AIDC system and the challenges in achieving intelligent lifecycle management.

\subsection{AIDC System Overview}
AIDCs are highly integrated, complex, and multidisciplinary cyber-physical systems (CPS). An AIDC comprises three major interconnected subsystems, i.e., the power supply system, the computing system, and the cooling system as shown in Fig.~\ref{fig:aidc}. Compared with traditional DC, the AIDC integrates innovative solutions to address the challenges posed by increasing power densities and the demand for sustainable operations. Specifically, the computing system in an AIDC is designed to deliver high-performance AI training and inference services while adhering to service level objectives (SLOs). To manage the thermal demands of high-power-density racks, the cooling system in an AIDC is equipped with liquid cooling technology. This advanced cooling method enables efficient heat removal directly at the source, reducing thermal resistance and ensuring optimal operating temperatures for IT equipment. Unlike traditional air-cooled systems, liquid cooling systems adopt cooling distribution units (CDUs) for distributing the coolant to various IT components in the data hall. The heated liquid is then transported to a heat exchanger, where it releases the heat to an external cooling plant. The power supply system in an AIDC is responsible for providing reliable electricity from multiple sources, including the power grid and on-site renewable energy generators. It is supported by an uninterruptible power supply (UPS) for backup during power failures and power distribution units (PDUs) to efficiently distribute power to IT devices and cooling facilities. The integration of these subsystems involves the interplay of various physical processes, such as fluid dynamics, heat transfer, and electromagnetism.

\subsection{Challenges of AIDC Digital Transformation}
Developing the AIDC digital twins is important to enable proactive and intelligent operations. It not only provides a large volume of synthetic data for various what-if analyses and ML training for offline use but also facilitates online optimization to achieve certain objectives (e.g., reducing cooling system power usage) subject to the physical constraints~\cite{wang2022toward}. However, the above physical processes are interrelated and occur at different scales, which presents several modeling challenges:

{\bf Large-Scale Layout Modeling}: When developing an AIDC digital twin, we need to create the 3D scenes of the physical layout and specify the physical properties of each equipment, such as the rated power of the servers, the setpoint of CRAHs, the liquid flow rate of CDU, etc. For an industry-grade AIDC with thousands of facilities and equipment, manually editing the model configuration file is error-prone and labor-intensive. This editing process becomes even more complex as AIDCs evolve dynamically, which requires continuous updates to reflect hardware upgrades or layout changes. Additionally, integrating the layout configuration with existing asset management systems to automate configuration is often hindered by proprietary formats or missing legacy infrastructure information, which further complicates the interoperability.  

{\bf Data Availability and Quality}: Accurate mechanistic AIDC modeling requires large amounts of high-quality data, which might not always be available. For example, it is difficult to collect dynamic temperature data in a data hall served by a stably operated cooling system. Physically collecting such data may incur high costs and potential risks. Moreover, the substantial expense of data collection and annotation can negatively impact the environmental sustainability of machine learning. Additionally, sensor networks may suffer from noise, faults, or sparse coverage, leading to gaps in spatial or temporal resolution and affecting data quality. Although synthetic data can supplement real-world datasets, validating its fidelity to actual scenarios remains challenging, especially when deployed sensors are sparse or simulation configurations are not calibrated~\cite{wang2020kalibre}.

{\bf Multiphysics \& Multiscale Nature}: The multiphysics and multiscale nature of AIDC demands interdisciplinary expertise for the mechanistic modeling. To illustrate, the heat transfer and airflow dynamics interact across scales from chip-level thermal fluctuations to room-level airflow patterns. These interactions are computationally intensive to simulate, particularly when using high-fidelity models like computational fluid dynamics (CFD). In addition, the temporal-spatial mismatches, such as rapid server workload changes versus slower temperature variations, require specially designed multiscale models. The details may be lost when simplifying the microscale phenomena (e.g., conjugate heat transfer on a liquid plate) to reduce computational costs. Therefore, balancing granularity with efficiency is essential for real-time optimization but remains a persistent challenge.

\section{Intelligence Evolution for DC Management}
\begin{figure*}[t]
    \centering
    \includegraphics[width=0.95\linewidth]{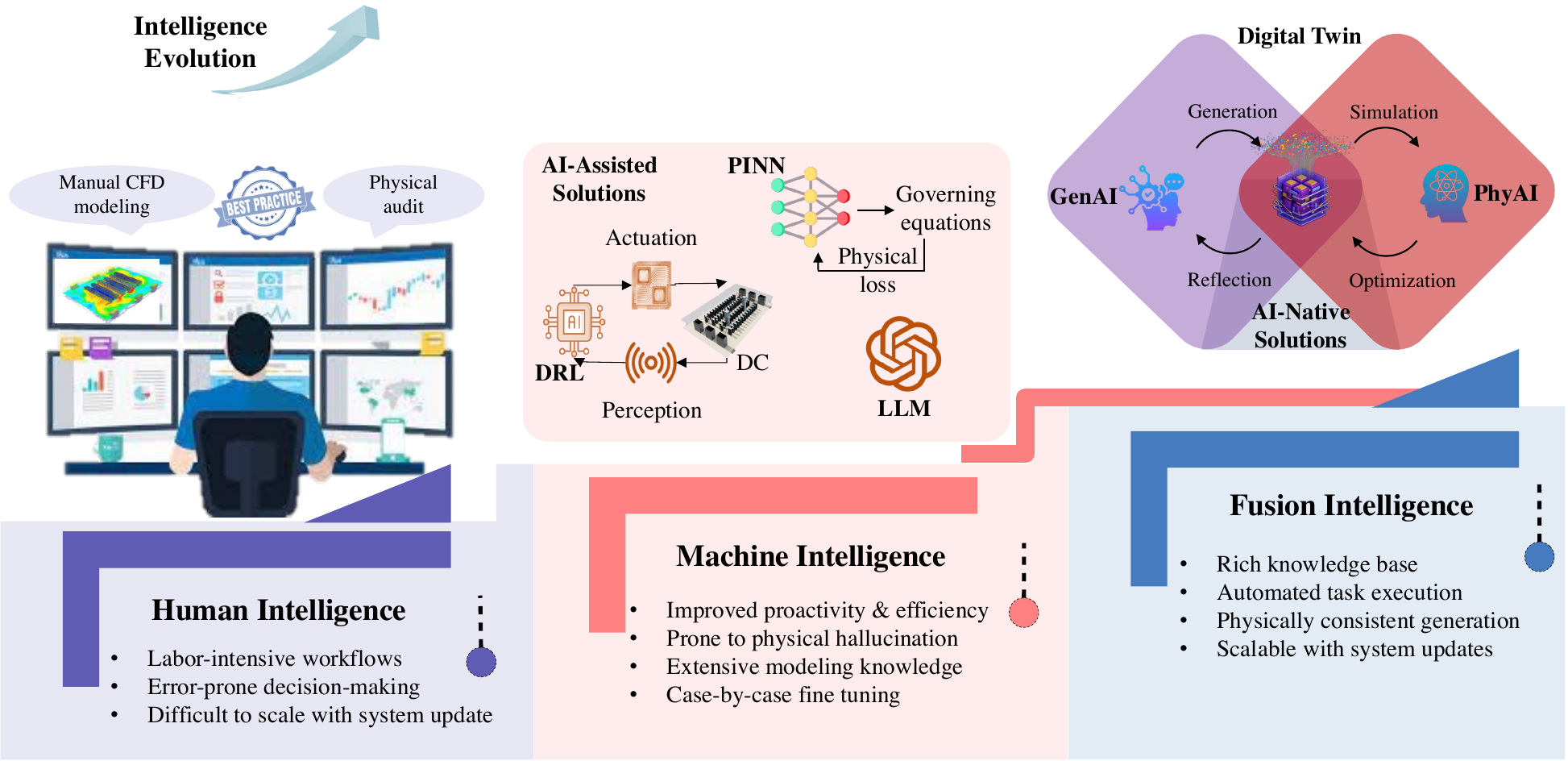}
    \caption{Intelligence evolution for DC management. Human intelligence relies heavily on human expertise and manual workflows; Machine intelligence introduces efficiency gains but faces challenges in generalizability and physical consistency; Fusion intelligence integrates GenAI’s automation with PhyAI’s physical grounding, creating a closed-loop framework for scalable and reliable DC operations.}
    \label{fig:intelligence}
\end{figure*}

This section reviews the evolution of DC management, from human intelligence to machine intelligence, and ultimately to fusion intelligence, as shown in Fig.~\ref{fig:intelligence}, with fusion intelligence proposed as a promising AI-native solution for AIDC digital transformation.

\subsection{Human Intelligence: The Era of Best Practices}
Human intelligence forms the foundation of early DC lifecycle management, which relies mostly on manual workflows and domain expertise. This level of intelligence leverages simulation and integrated platforms to support designers' and operators' best practices. For example, cooling system design depends on iterative simulations (e.g., CFD tools) and physical audits to mitigate potential risks such as hotspots. The Data Center Infrastructure Management (DCIM) system enables operators to monitor system status and take appropriate actions based on sensor data. While these tools provide valuable insights for responding to abnormalities and failures, they are labor-intensive, error-prone, and lack scalability, making it challenging to keep pace with increasing infrastructure complexity and dynamic workloads. As systems become more complex, advancing DC management with automated and proactive capabilities becomes essential.

\subsection{Machine Intelligence: The Rise of AI-Driven Automation}
Machine intelligence introduces automated and proactive management through AI-driven solutions such as GenAI and PhyAI. On the one hand, GenAI has emerged as a promising tool for automating workflow execution, data analysis, and scene design. For example, LLMs have been proposed to automate digital scene generation due to their remarkable natural language understanding and reasoning capabilities~\cite{feng2023layoutgpt}. However, GenAI may produce hallucinations when applied to AIDC. In AIDC applications such as layout design, GenAI may generate physically infeasible layouts due to a lack of constraint enforcement (e.g., design standards, thermal balance). These hallucinated designs then require costly manual corrections, compromising the automation advantages of GenAI. On the other hand, PhyAI has been proposed as a promising solution for developing real-time, high-fidelity mechanistic models~\cite{karniadakis2021physics} and intelligent policies by integrating physical laws through neural network architectural design~\cite{wang2020kalibre}, loss function regularization~\cite{wang2023phyllis}, or inference process constraints~\cite{cao2022reducio}. These physics-informed models can then be used for various what-if analyses and operational optimizations. For example, these models can serve as a virtual environment to enable DRL-based agent interactions for developing energy-efficient cooling control policies~\cite{wang2023phyllis}. While PhyAI reduces modeling data demands and improves physical plausibility, it requires extensive domain expertise for specific knowledge injection. Despite efficiency gains, existing siloed machine intelligence fails to unify domain knowledge, automation, and physical consistency, limiting the scalability of such intelligent solutions for wide deployment. Therefore, it is desirable to integrate the advantages of different intelligent solutions into a scalable framework.

\subsection{Fusion Intelligence: The Path to AI-Native Solutions}
While GenAI and PhyAI individually offer distinct advantages in model generation and optimization, their respective limitations, i.e., GenAI’s hallucinations and PhyAI’s scalability gaps, underscore the need for a unified framework that synergizes their strengths to compensate for their weaknesses. This synergy aims to bridge the gap by integrating GenAI’s automation with PhyAI’s domain optimization in a collaborative manner. Such collaboration provides an AI-native solution for MCI digital transformation. For example, GenAI can translate natural language goals and textual descriptions into structured digital scene description files, codes of model structures, variables, or relationships, while PhyAI validates and optimizes specific parameters of the generated structures based on loss functions defined by the conservation of physical laws and real-time data deviations. While a few recent studies~\cite{ma2024llm,holtautomatically} have demonstrated the efficacy of LLMs for mechanistic model generation, they do not address the challenges of lifecycle management of complex infrastructures like AIDCs. Accordingly, this paper aims to propose a fusion intelligence framework that integrates GenAI and PhyAI for large-scale AIDC lifecycle management and highlights its potential for a broad range of MCI applications. The proposed framework is expected to establish a new paradigm for robust, scalable, and intelligent solutions to accelerate MCI's digital transformation.

\section{The Fusion Intelligence Framework and Relevant Applications}
This section proposes the fusion intelligence framework and formalizes the collaboration mechanism.

\subsection{Definition of Agents and Their Fusion}
\begin{figure}[t]
    \centering
    \includegraphics[width=1\linewidth]{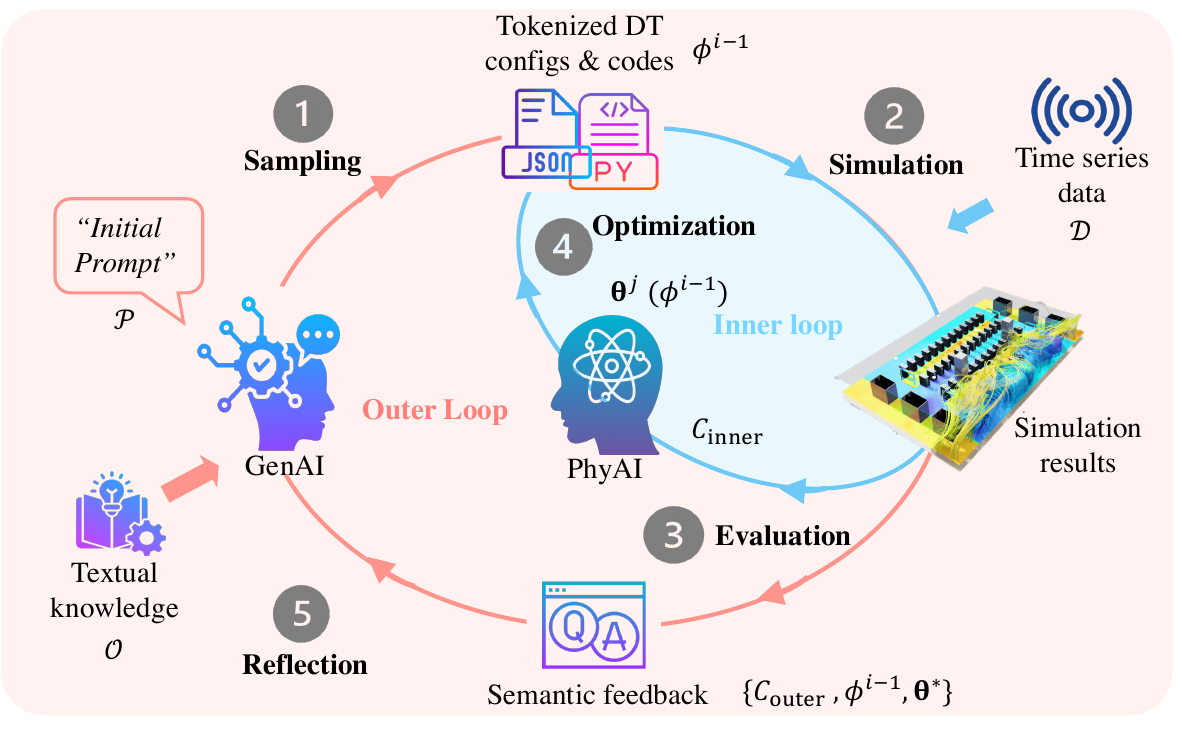}
    \caption{The fusion intelligence operates as follows: 1) The GenAI samples tokenized digital twin configurations and codes from prompts and the feedback from the PhyAI optimization. 2) These samples are then simulated using a time series dataset. 3) The simulation results are evaluated and organized into numerical errors and semantic descriptions, providing feedback for PhyAI and GenAI, respectively. 4) The PhyAI optimizes model parameters and 5) GenAI reflects the semantic results to refine the generation. 
    }
    \label{fig:framework}
\end{figure}

We propose the fusion intelligence framework that fuses GenAI and PhyAI into a dual-agent collaboration, as illustrated in Fig.~\ref{fig:framework}. To formalize the concept, we first define the individual agents and the fusion of their capabilities for twinning the AIDCs.

\begin{definition}[Agents]
In this framework, the GenAI and PhyAI agents act as two distinct agents. GenAI leverages its reasoning capabilities to generate and refine tokenized digital twins, while PhyAI utilizes its domain grounding to optimize the parameters associated with the digital twins.
\end{definition}

\begin{definition}[Fusion Intelligence]
Fusion intelligence integrates GenAI and PhyAI into a dual-agent collaborative mechanism for the automated and physics-compliant creation and refinement of digital twins within a closed-loop system.
\end{definition}

% shows the proposed fusion intelligence framework, which integrates GenAI and PhyAI into a closed-loop dual-agent collaborative system for automated, physics-compliant AIDC digital twin development. We illustrate the collaboration as below: With the above tri-level fusions, the proposed framework is expected to provide an AI-native solution to advance MCI digital transformation.

\subsection{Formulation of GenAI-PhyAI Collaboration}
We can analyze the interaction between GenAI and PhyAI within the fusion intelligence framework through a hierarchical optimization. In this context, GenAI and PhyAI act as two distinct agents collaborating towards the goal of creating and refining an optimal digital twin.
Formally, we consider a set of time series data \(\mathcal{D} = \{(\mathbf{s}_t, \mathbf{a}_t)|t\in[T]\}\) collected at discrete time steps \(t=0, 1, ..., T\) where \(\mathbf{s}\) is a vector of system states and \(\mathbf{a}\) is a vector of control actions. A digital twin, denoted by \(f_\text{DT}\), aims to predict the next system state given the current state-action pair by \(\mathbf{\hat{s}}_{t+1} = f_\text{DT}\left((\mathbf{s}_t, \mathbf{a}_t) | \phi, \uptheta \right)\), where \(\phi \in \mathcal{T}\) represents the semantic DT configurations (e.g., JSON/XML-defined 3D layouts and Python-based mechanistic models) that encodes the model architecture or physical equations, \(\uptheta(\phi) \in \Theta\) represents a vector of parameters associated with the structure (e.g., 3D scene coordinates or physical properties), \(\mathcal{T}\) and \(\Theta\) are the possible digital twin structure and parameter space, respectively. %The collaboration can be modeled as a sequential game, specifically resembling a Stackelberg game, where the 
In this collaboration, GenAI acts as the leader by proposing a semantic structure \(\phi\), while PhyAI acts as the follower, responding by finding the optimal parameters \(\uptheta\) for that specific structure. We next illustrate the collaboration strategy and corresponding objective functions.

 %\mathcal{D} \(\mathcal{D}\) is a set of sensor-measured data that can be used to update \(\uptheta\) dynamically
 
% $\blacksquare$ 
% \subsection{Strategies \& Objective Functions}

\subsubsection{Outer Loop Generative Refinement}
This GenAI agent represents the decision-making process of the leader as shown in the outer loop of Fig.~\ref{fig:framework}. It aims to propose the semantic structure, \(\phi\), of the digital twin. This process is iterative and leverages GenAI's self-reflection ability to refine its proposals over time. The loop starts with an initial natural language prompt \(\mathcal{P}_\text{init}\) from a human user outlining the objectives and constraints for the digital twin (e.g., ``Design an AIDC with 10PFLOPS compute and PUE under 1.3"). In subsequent iterations \(i>0\), the GenAI receives feedback derived from the PhyAI agent's optimization results from the previous iteration \(i-1\). This feedback includes the previously proposed structure, the PhyAI optimized parameters \(\uptheta^*\) and
a cost function, denoted by \(\mathcal{C}_\text{outer}\), that evaluates the overall quality, adherence to user intent, and performance of the optimized digital twin. At each new step, the GenAI leverages its pre-trained knowledge and the provided in-context examples from the prompt and feedback to sample one or more candidate structures as:
\begin{equation}
    \phi^{(i)} = \text{GenAI}\left(\{\mathcal{C}_\text{outer}(f_\text{DT})_k, \phi^{(i-1)}_k, \uptheta^*_k\} \, \big| \, \mathcal{P}_\text{init}, \mathcal{O}\right),
\end{equation}
where \(k \in [K]\) is the top-\(K\) results based on \(\mathcal{C}_\text{outer}\). In this loop, the GenAI's self-reflection optimization can be triggered via mechanisms like in-context learning or reinforcement learning.

% Through in-context learning mechanism, the GenAI refines its strategy that aims to minimize the cost function by:
% \begin{equation}
% \phi^* \triangleq\ \operatorname*{arg\,min}_{\phi \in \mathcal{T}} \quad \mathcal{C}_\text{outer}\left(f_\text{DT}\left(\phi, \uptheta^*(\phi)\right)\right). 
% \end{equation}
% In this loop, the GenAI's self-reflection optimization can be triggered via mechanisms like in-context learning or reinforcement learning rewards.

% $\blacksquare$ 
\subsubsection{Inner Loop Physics-Driven Optimization}  
The PhyAI agent serves as the follower as illustrated in the inner loop of Fig.~\ref{fig:framework}. Given a semantic structure \(\phi\) proposed by the GenAI leader, the PhyAI follower seeks to determine the optimal set of parameters \(\uptheta(\phi)\) for that structure using gradient descent optimization, grounding the results in physical laws and real-world data. For example, if \(\phi\) defines a neural network representing a heat exchanger, \(\uptheta(\phi)\) would correspond to the network’s weights and biases. If \(\phi\) is a 3D layout, \(\uptheta(\phi)\) might include precise coordinates or physical properties of the equipment. The optimized results are then structured as semantic feedback for GenAI to further refine generation.

The overall interaction is fundamentally cooperative, as both agents work towards the shared system-level goal of generating an accurate, physically plausible, and optimized digital twin that satisfies the user's requirements. This interaction can be formalized as a bi-level optimization problem:
\begin{equation}
    \begin{aligned}
& \operatorname*{min}_{\phi \in \mathcal{T}} \quad \mathcal{C}_\text{outer}\left(f_\text{DT}\left(\phi, \uptheta^*(\phi)\right)\right)  \quad \quad \textit{(Outer Loop)} \\
\text{s.t.} &\quad \uptheta^* \in \operatorname*{arg\,min}_{\uptheta \in \Theta} \quad \mathcal{C}_\text{inner}\left(f_\text{DT}\left(\phi, \uptheta(\phi)\right)\right) \quad \quad \textit{(Inner Loop)},
\end{aligned}
\end{equation}
where \(\mathcal{C}_\text{outer}\) can be evaluated using the validation dataset while \(\mathcal{C}_\text{inner}\) is evaluated using the training dataset. The proposed fusion intelligence framework is expected to provide an AI-native solution to accelerate AIDC digital transformation. 

% three-stage workflow:
% {\tiny\encircle{\normalsize1}} GenAI agent generates tokenized structures (e.g., JSON/XML-defined layouts, Python-based mechanistic models) from natural language prompts and a SimReady device library.
% {\tiny\encircle{\normalsize2}} These structures are then simulated and evaluated using metrics such as temperature distribution, PUE, or mean absolute errors (MAEs) with real-time data. The evaluated results are organized into numerical errors and semantic descriptions, serving as feedback for PhyAI and GenAI, respectively.
% {\tiny\encircle{\normalsize3}} PhyAI agent optimizes model parameters based on domain constraints and prediction errors, while deviations trigger GenAI self-reflection through in-context learning. The dual-agent collaboration can be illustrated as follows

% This interaction is formalized as a bi-level optimization problem:
% \begin{equation}
%     \begin{aligned}
% &\min_{\phi \in \mathcal{T}} \quad \mathcal{J}_\text{eval}\left(\phi, \uptheta^*(\phi)\right) \quad \quad \textit{(Outer Loop)} \\
% &\text{s.t.} \quad \uptheta^*(\phi) \in \arg\min_{\uptheta \in \mathbb{R}^n} \mathcal{J}_\text{piml}(\phi, \uptheta) \quad \textit{(Inner Loop)},
% \end{aligned}
% \end{equation}
% where \(\phi \in \mathcal{T}\) is the tokenized structure (e.g., JSON, python code) generated by the GenAI, \(\mathcal{T}\) is the token space.    

% \subsection{Key Innovations}

\section{System Architecture \& Relevant Applications}
This section presents a system architecture that integrates the fusion intelligence framework for AIDC-related lifecycle management and applications.

\begin{figure}[t]
    \centering
    \includegraphics[width=1\linewidth]{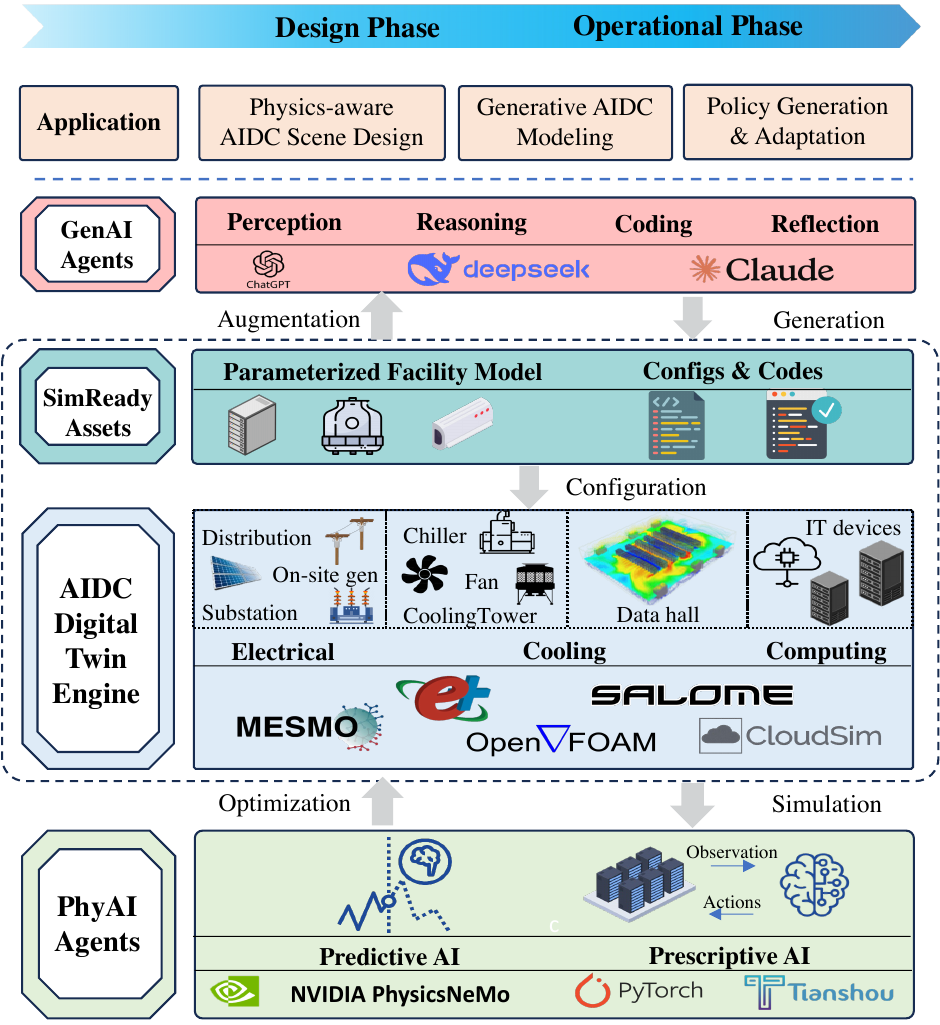}
    \caption{The proposed system architecture integrates the capabilities of GenAI and PhyAI agents into a fusion intelligence framework designed for AIDC lifecycle management and relevant applications.
    }
    \label{fig:system}
\end{figure}

\subsection{System Architecture \& Core Components}
To fuse the capabilities of GenAI and PhyAI for AIDC lifecycle management, we develop an integrated system as shown in Fig.~\ref{fig:system}. The system consists of the following components:
\begin{itemize}
    \item {\bf GenAI Agents}: The GenAI agents are designed to facilitate digital twin design and generation based on user-input natural language prompts. Specifically, we develop four types of agents, i.e., perception, reasoning, coding, and reflection. Each agent performs distinct task based on their defined functionalities. To achieve this, we adopt open-source LLMs such as OpenAI-o3-mini and DeepSeek-R1-32B as the foundation model to develop the GenAI agents. These models possess natural language understanding and reasoning capabilities, allowing them to extract user objectives and generate desired digital twin configurations, such as JSON-based scene description files or Python-based mechanistic models.
    \item {\bf SimReady Assets \& Digital Twin Engine}: The SimReady assets library contains geometric features, physical properties, and functional parameters of AIDC equipment such as liquid-cooled servers, CDUs, CRAHs, chillers, cooling towers, etc. It enables semantic search and parameter inference via a knowledge graph. This library serves as a knowledge base for the GenAI. In addition, it also contains the necessary configuration files or codes to run a digital twin simulation. The digital twin simulation aims to validate the generated design and provide synthetic data for PhyAI optimization. Specifically, this engine consists of several open-source software, such as OpenFOAM and EnergyPlus, to model the electric-cooling-computing coordinated AIDC system. 
    \item {\bf PhyAI Agents}: The PhyAI agents primarily consist of two types of AI agents: predictive AI and prescriptive AI. Specifically, the predictive AI aims to enhance digital twin prediction fidelity by optimizing or calibrating parameters, while the prescriptive AI is designed to learn intelligent policies for safe and efficient AIDC operations using data from the high-fidelity digital twin system. We develop the PhyAI agents based on NVIDIA's PhysicsNeMo~\cite{physicsnemo} and Tianshou~\cite{weng2022tianshou}. The PhysicsNeMo library is designed for physics-informed AI model training and inference. It includes various AI models for solving a range of forward and inverse problems. Tianshou is a state-of-the-art reinforcement learning library used for developing intelligent prescriptive AI policies.
\end{itemize}
With these components, we next illustrate the potential of the proposed fusion intelligence framework for advancing AIDC lifecycle management with relevant applications.

\subsection{Applications for AIDC Lifecycle Management}
The AIDC lifecycle spans from design to operations. The proposed fusion intelligence can empower intelligent digital transformation at different stages.
 
\subsubsection{Physics-Compliant Automated AIDC Design}
Traditional AIDC design is a multi-step, iterative process that blends human expertise with specialized tools to address a wide array of technical and regulatory considerations, such as security, space availability, energy efficiency, government regulations, workforce skills, and environmental hazards. Specifically, traditional data center (DC) design begins by gathering detailed requirements and selecting an appropriate site. Designers then create an initial layout with critical infrastructure and conduct simulations to validate performance metrics such as PUE or thermal safety~\cite{cadence}. The design is iteratively refined by humans to meet industry standards, which is a time-consuming and labor-intensive process. For instance, designing an industry-grade data hall requires collecting a vast amount of information and configuring thousands of geometric or thermodynamic properties for simulation, typically consuming several weeks~\cite{karniadakis2021physics}.

Fusion intelligence revolutionizes this workflow through automated design synthesis, multi-physics coupling simulation, and closed-loop refinement. It provides a new paradigm for text-to-design generation and optimization. Specifically, users can articulate design requirements in natural language and input them into a GenAI agent, like a LLM. The GenAI then parses natural language requirements (e.g., ``Design a 10MW AIDC with liquid cooling and PUE less than 1.3") into the structured template, like JSON/XML file, based on a SimReady library of pre-validated equipment and external knowledge base of site information. Next, a PhyAI agent, such as a PINN model, simulates the generated design and optimizes key parameters through automatic differentiation. Finally, the simulation results are evaluated and fed back to the GenAI for reflective optimization based on in-context learning until it satisfies the design objectives. This method enhances the physical compliance of the GenAI's generated design through closed-loop optimization with physical feedback.

\subsubsection{Generative Multi-Scale Mechanistic Modeling}
Traditional mechanistic modeling requires extensive domain knowledge to build the physical geometry and program physics-based equations. To solve the equations, numerical algorithms like finite volume methods are often adopted. However, these methods are computationally expensive and difficult to adapt as system complexity increases. PhyAI accelerates inference through a neural network forward process, but training PhyAI typically requires domain expertise to curate the model structure and loss functions. For AIDC with heterogeneous coupling systems, the model structure and loss functions must be carefully designed. For example, developing air-liquid hybrid cooling mechanistic models requires consideration of both macroscopic phenomena, such as airflow at the room level, and microscopic phenomena, such as heat transfer at the chip level within individual servers. 

The fusion intelligence framework revolutionizes mechanistic modeling by integrating automated code generation, physics-informed optimization, and iterative refinement into a unified, closed-loop workflow. This approach establishes a new paradigm for text-to-model generation, enabling the rapid development of high-fidelity, domain-specific digital twins for complex systems AIDC systems. Specifically, the GenAI acts as a ``digital engineer" by
leveraging its pretrained knowledge of physics and programming syntax to translate facility specifications (e.g., the heat exchanger of a liquid-cooled AIDC) into executable code. For example, the GenAI can generate neural operator networks for room-scale airflow and PINNs for server-level heat transfer. PhyAI integrates real-time sensor data, simulation data, and the generated physical constraints to update model parameters, enabling efficient model updates and online data assimilation. The generated model architecture and optimization results are then fed back to the GenAI for possible improvement.

\section{Case Study}
To demonstrate the performance of the proposed fusion intelligence framework, this section presents two case studies on physics-aware equipment selection for AIDC design and generative modeling for a heat exchanger. 
\begin{figure}[t]
    \centering
    \includegraphics[width=1\linewidth]{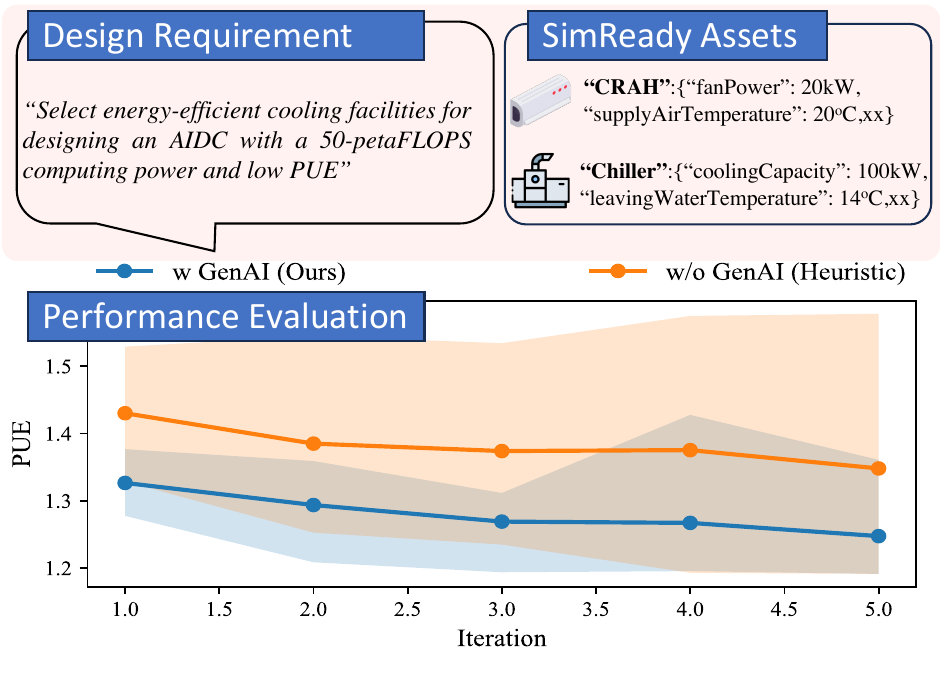}
    \caption{Case 1: Evaluation of GenAI in physics-aware equipment selection. We can observe the introduction of GenAI accelerates design PUE improvement.}
    \vspace{-1em}
    \label{fig:case1}
\end{figure}

\subsection{Case One: Physics-Aware AIDC Equipment Selection}

\subsubsection{Case Setup}
In this case, our goal is to select proper equipment that meets user-defined objectives at the AIDC design stage. The purpose is to demonstrate the effectiveness of introducing GenAI and physical simulation for equipment selection. The selection criteria should incorporate several key factors, including computing performance, energy efficiency, and adherence to facility layout standards. The generated design should include the optimal combination of facilities and their spatial relationships. Specifically, users will input prompts such as: ``{\em I would like to design an AIDC in a tropical region with a computing performance of 50-petaFLOPS and energy-efficient design. Please assist in selecting appropriate facilities and generating a data hall layout to achieve this goal}." In response, the GenAI will produce semantic configurations that define facility relationships and properties for physical simulation. These generated configurations are optimized to meet specified constraints, including geometrical rules outlined in~\cite{li2023chattwin}. Simulation results are then fed back into the GenAI system to iteratively refine the designs. We use OpenAI-o3-mini as the GenAI model to interpret user prompts and select appropriate components from an in-house device library containing over 1,000 AIDC facilities, along with historical weather data from over 100 locations. EnergyPlus is used to calculate energy efficiency metrics such as PUE to inform the feedback process.

\subsubsection{Evaluation Results}
To demonstrate the effectiveness of the framework, we compare the PUE optimization results with those of a heuristic evolutionary algorithm that selects equipment by applying mutation and selection to an initial set of randomly sampled candidates, guided by simulation feedback. Figure~\ref{fig:case1} presents the PUE variation results over five outer loop iterations. For each iteration, we plot the results of the top five candidates and use them as feedback for the LLM. The dot values represent the averages of five independent experiments, and the shaded areas indicate the standard deviations. From the results, we observe that with the introduction of the LLM, the design is initialized with better performance, and the PUE is optimized to 1.25. In contrast, without the LLM, the PUE can only be optimized to 1.35. This improvement is attributed to the LLM’s remarkable reasoning capabilities, which enable it to interpret design requirements and select suitable equipment from the knowledge base. This case study demonstrates the effectiveness of integrating GenAI with physical simulators for AIDC design.

\begin{figure}[t]
    \centering
    \includegraphics[width=1\linewidth]{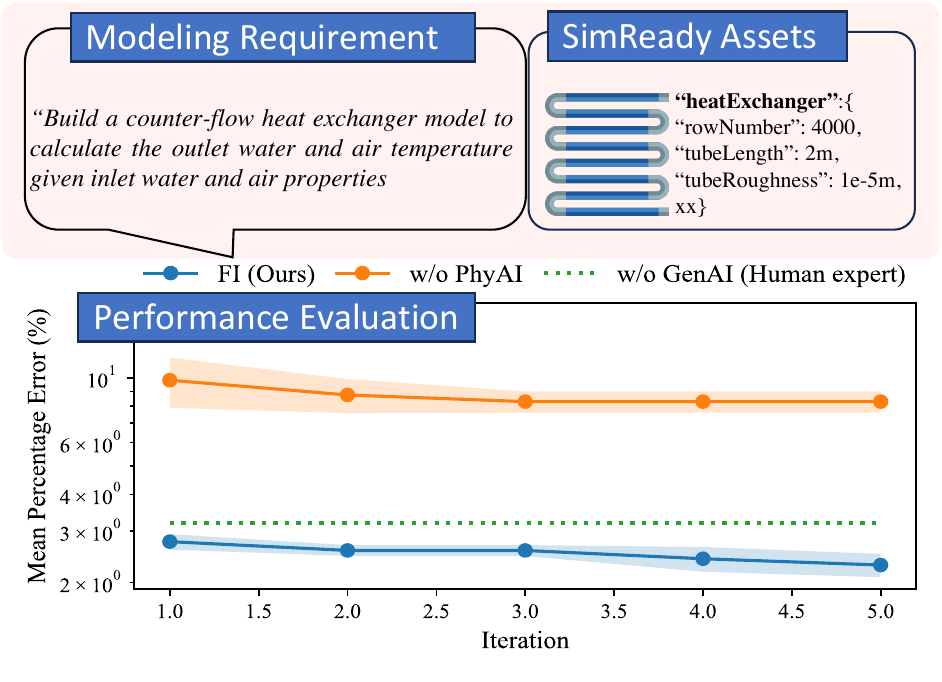}
    \caption{Case 2: Evaluation of fusion intelligence for heat exchanger modeling. The proposed framework accelerates model prediction accuracy improvement.}
    \vspace{-1em}
    \label{fig:case2}
\end{figure}
% We compare the fusion intelligence solution with the following baselines for the first case. The baselines include i) {\bf genetic algorithm}, which generates designs by applying mutation and selection to an initial set of randomly sampled candidates, guided by simulation feedback; ii) {\bf LLM w/o reflection}, which relies solely on the LLM for design generation and disables the feedback loop from physical simulation to the LLM.

% \subsubsection{Evaluation Results} Fig.~\ref{fig:results}(a) shows the evaluation results for physics-aware scene generation. Fusion intelligence reduces design time from 72 hours (manual) to 2 hours, representing a 97\% improvement. Additionally, the proposed fusion framework achieves a PUE of 1.28 after 5 generation-reflection iterations, outperforming manual designs (1.35) and heuristic methods (1.32). Layouts without feedback fail to meet PUE targets, resulting in a PUE of 1.45. 

\subsection{Case Two: Generative Modeling for a Heat Exchanger}

\subsubsection{Case Setup}
In this case, our goal is to generate a mechanistic model of a heat exchanger in Python, denoted by \(\Phi_\text{hx}\), to simulate heat transfer between different mediums. The purpose is to demonstrate the effectiveness of fusing GenAI and PhyAI for mechanistic modeling. This model characterizes the physical phenomena of CRAH coils or CDU heat exchangers, aiming to predict the air and water outlet conditions based on their inlet properties, such as temperature and mass flow rate. Due to commercial interests, complete and detailed information about the heat exchanger, such as tube length, diameter, and thickness, is unavailable. We treat these as unknown parameters, \(\Theta_\text{hx}\), to be learned from sensor-measured data. Similar to the first case, we use OpenAI-o3-mini to generate the code structure based on a model description such as: ``{\em Provide me with a chilled water coil model built as counter-flow. The model should be able to calculate the outlet water temperature and outlet air temperature, given the inlet/outlet water or air flow rate and temperature. The model should expose several parameters that can be optimized to improve performance}."
The parameters are learned from data collected in a production data center, as illustrated in~\cite{cao2025adaptive}.

\subsubsection{Evaluation Results}
We choose the mean percentage error (MPE) as the metric to evaluate the performance of the generated model in contrast to the one developed by an expert in~\cite{cao2025adaptive}. Figure~\ref{fig:case2} shows the MPE variation. Similar to the first case, we run multiple independent experiments and plot the results of the top five candidates for each iteration. The dotted orange line indicates the performance of the pure-physics model developed by the human expert without the introduction of the LLM. We also compare the performance with models generated purely by LLMs, without PhyAI’s inner-loop parameter optimization. However, these LLM-generated models exhibit significantly higher errors up to 80\%. To maintain conciseness, we omit these results from the figures. From the results, we observe that with the collaboration of PhyAI and GenAI, the MPE of 2.2\% is significantly lower than the 6.3\% achieved by the expert-developed model. This improvement achieved by fusion intelligence can be attributed to the LLM’s reasoning capabilities and rich domain knowledge. Upon inspecting the generated model code, we find that the LLM uses logarithmic re-parameterization for parameters across different ranges and incorporates a residual network to augment the original white-box models. These design choices enhance training convergence and compensate for modeling errors introduced by incomplete physical knowledge.

% the following baselines for the second case. The baselines include i) {\bf Mechanistic model w/o GenAI}, which involves developing the heat exchanger mechanistic model by a heat transfer expert using thermodynamic principles and a set of psychrometric functions; {\bf Mechanistic model w/o PhyAI}, which is built based on textual descriptions but does not incorporate parameter learning from sensor-observed data.

% \subsubsection{Evaluation Results}
% Fig.~\ref{fig:results}(b) shows the generated mechanistic model accuracy. Fusion PINNs achieve prediction accuracy higher than 95\%, outperforming pure data-driven models (80\%). In addition, fusion reduces design iterations from 10+ (physics-based) to 3, slashing computational costs by 70\%.

\section{Future Directions}
In this section, we envision several research directions of the fusion intelligence framework for future investigation.

\subsection{Hybrid Feedback Mechanisms for GenAI Refinement}
The feedback for GenAI reflection is important to optimize the generation for the desired objective. Our case study at this stage only uses semantic information to guide the generation via in-context learning.
Future feedback mechanism design could involve a hybrid approach combining RL rewards with semantic feedback. For example, PhyAI can provide model performance metrics as the RL reward, which the GenAI uses to adjust its own weights. Also, iterative refinement loops where the GenAI generates multiple candidates, and PhyAI selects the best based on multi-objective criteria. This approach would bridge numerical errors with actionable textual guidance, enabling iterative refinement of JSON layouts or Python-based models while maintaining explainability.

\subsection{Prompt Design with Physics-Aware Structured Templates}
Future work could focus on designing physics-structured prompts that embed domain-specific constraints (e.g., thermal limits, airflow equations) directly into GenAI instructions. For example, prompts could dynamically integrate parameters from the simulation-ready library (e.g., server power curves, liquid cooling plate dimensions) using retrieval-augmented generation (RAG), reducing hallucinations by grounding outputs in verified physical properties. Multi-step prompting strategies, such as decomposing user queries into layout, thermodynamics, and compliance sub-tasks, could further align GenAI outputs with engineering standards, enabling automated generation of objective-aware designs.

\subsection{Few-Shot Automated Modeling for Novel Devices}
To adapt to emerging hardware (e.g., quantum cooling systems), few-shot knowledge injection could enable GenAI to infer device parameters from minimal data (e.g., spec sheets) using contrastive learning. PhyAI could then fine-tune generated models via transfer learning, reusing pre-trained neural operator architectures for airflow or heat transfer while updating only critical layers. For example, a new liquid-cooled GPU rack’s thermal dynamics could be modeled by extending PINNs with a modular microchannel heat transfer subnetwork, initialized using GenAI-parsed vendor documentation.

\subsection{Dynamic Data Assimilation with Adaptive Loss Weighting}
Online data assimilation needs mechanisms to efficiently handle streaming data and dynamically adjust the weights of the loss function. On the one hand, a sliding-window attention mechanism can be applied to prioritize real-time sensor streams (e.g., temperature, flow rates) over historical data. On the other hand,  the loss function of the mechanistic model is a weighted sum of multiple terms, including the mismatch of observed online data, physical residuals, and boundary constraints in the form of $\mathcal{L} = \lambda_1 \mathcal{L}_{\text{data}} + \lambda_2 \mathcal{L}_{\text{phys}} + \lambda_3 \mathcal{L}_{\text{boundary}}$ where $\lambda_1$, $\lambda_2$ and $\lambda_3$ are weights of different terms. Adaptively adjusting these weights is important to balance the model's data interpolation fidelity and physical plausibility.

\section{Conclusion}
In this article, we propose the fusion intelligence, a transformative framework that harmonizes the generality of GenAI with the specificity of PhyAI through a closed-loop, dual-agent collaboration mechanism. By integrating GenAI’s ability to interpret natural language prompts and generate tokenized digital scenes with PhyAI’s capacity to enforce domain constraints, the framework addresses critical challenges such as thermal risks, scalability bottlenecks, and cross-domain expertise shortages. Case studies demonstrate its efficacy in energy-efficient AIDC design and mechanistic modeling. Fusion intelligence provides a blueprint for managing complex MCIs across sectors toward an autonomous and optimized level.

% \input{acknoledgement}
% \balance
\bibliographystyle{IEEEtran}
\bibliography{ref}

\end{document}